\documentclass[sigconf]{acmart}
\AtBeginDocument{%
  }

\renewcommand\footnotetextcopyrightpermission[1]{}
\acmConference[]{}{}{}
\acmBooktitle{}

\begin{document}

\title[Vid-Freeze: Protecting Images from Malicious I2V Generation]{Vid-Freeze: Protecting Images from Malicious Image-to-Video Generation via Temporal Freezing}

\author{Rohit Chowdhury}
\authornote{Samsung Research Institute - Bangalore}
\authornote{Equal contribution}

\affiliation{%
  \country{}
}
\email{rohit.c@samsung.com}

\author{Aniruddha Bala}
\authornotemark[2]
\authornotemark[1]
\affiliation{%
  \country{}}
\email{aniruddha.b@samsung.com}

\author{Rohan Jaiwal}
\authornotemark[1]
\affiliation{%
  \country{}}
\email{r.jaiswal@samsung.com}

\author{Siddharth Roheda}
\authornotemark[1]
\affiliation{%
  \country{}}
\email{sid.roheda@samsung.com}

\renewcommand{\shortauthors}{Chowdhury et al.}

\begin{abstract}
  The rapid progress of image-to-video (I2V) generation models has introduced significant risks by enabling deceptive or malicious video synthesis from a single image. Prior defenses such as I2VGuard attempt to immunize images by inducing spatio-temporal degradation, which does not necessarily provide meaningful protection, since residual motion can still convey malicious intent. In this work, we introduce Vid-Freeze -- a novel adversarial defense that adds imperceptible perturbations to enforce temporal freezing in generated videos. Our method explicitly targets attention dynamics in I2V models to suppress motion synthesis. As a result, immunized images produce standstill or near-static videos, effectively blocking malicious content generation. Experiments demonstrate strong protection across models and support temporal freezing as a promising direction for proactive and meaningful defense against I2V misuse.

\end{abstract}



\keywords{Image-to-Video Generation , Adversarial Defense, Image Immunization , Temporal Freezing , Attention Suppression}
\begin{teaserfigure}
  \centering
  \includegraphics[width=0.9\textwidth]{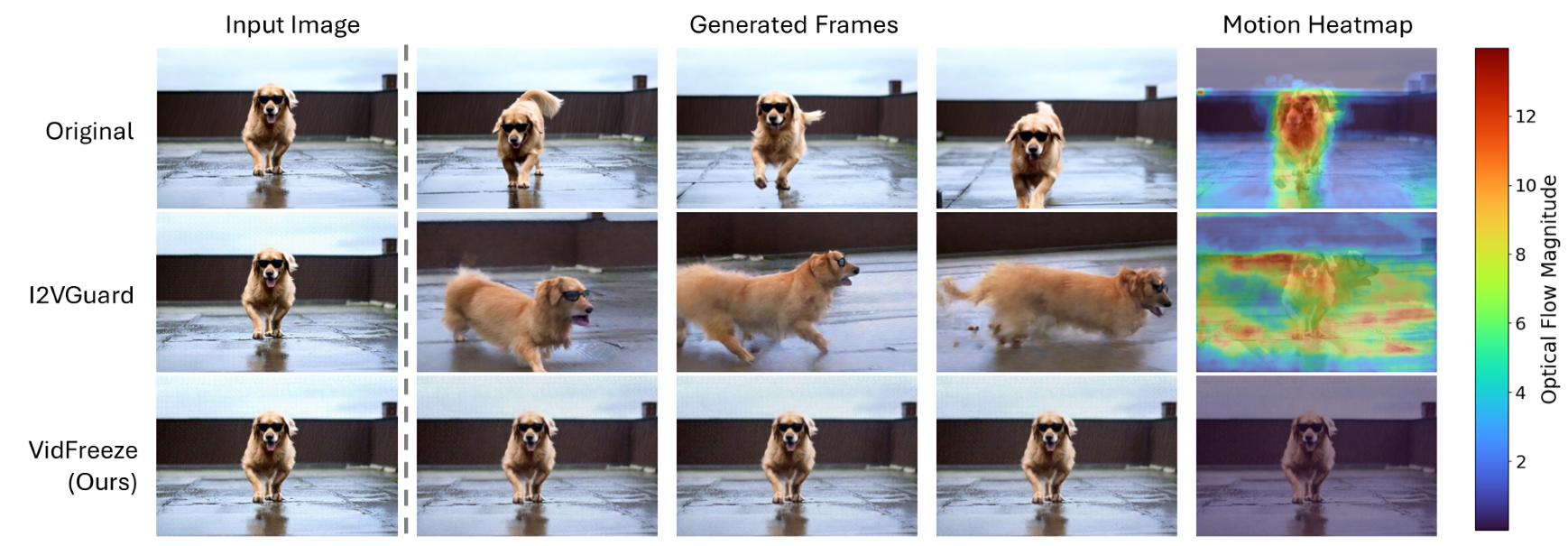}
  \caption{Vid-Freeze as a defense against malicious image-to-video generation. Rows show CogVideoX outputs for the original image and for images immunized with I2VGuard and Vid-Freeze, along with corresponding motion heatmaps. Vid-Freeze provides strong and meaningful protection by producing static (frozen) videos, whereas I2VGuard yields spatio-temporally degraded outputs that may still convey malicious intent. Prompt: "Dog running towards the camera".}
  \Description{Comparison of generated video frames and motion heatmaps for original, I2VGuard-immunized, and Vid-Freeze-immunized input images, where Vid-Freeze yields near-static outputs.}
  \label{fig:teaser}
\end{teaserfigure}


\maketitle

\section{Introduction}
\label{sec:intro}

The rise of diffusion-based generative models has accelerated progress in video synthesis, enabling image-to-video (I2V) systems that can transform static images into realistic videos while preserving the subject’s identity. Frameworks such as Stable Video Diffusion \cite{sd}, CogVideoX \cite{yang2024cogvideox}, AnimateDiff \cite{guo2024animatediff}, Animate-Anyone \cite{hu2023animateanyone}, and ControlNeXt \cite{peng2024controlnext} exemplify this progress, enabling controllable motion and semantic alignment for applications in entertainment, advertising, and virtual content creation. However, the same capabilities pose serious risks. Malicious users can exploit I2V models to fabricate deceptive or unauthorized videos, threatening privacy, security, and intellectual property.
Recent work such as I2VGuard \cite{gui2025i2vguard} attempts to disrupt spatial content and temporal consistency in generated videos, yet it falls short of fully blocking motion, allowing residual dynamics to persist. Consequently, videos generated from I2VGuard-immunized images do not necessarily provide meaningful protection: despite visible spatio-temporal disruption, the remaining motion can still carry the semantic intent of a malicious prompt (see Fig.~\ref{fig:teaser}). In contrast, we argue that achieving complete temporal freezing of the generated video, such that only the input frame is reproduced across all timesteps, represents a far more robust form of protection against image-to-video misuse. To this end, our contributions are threefold:

(1) we propose \texttt{Vid-Freeze}, a targeted adversarial defense that forces image-to-video models to generate static or temporally frozen videos;
(2) we introduce a novel freezing loss with target attention that effectively suppresses motion, collapsing generated videos toward the input image; and
(3) we show that temporal freezing provides stronger protection than spatio-temporal degradation methods such as I2VGuard, with detailed empirical analyses, visualizations and ablations to support this finding.

\section{Related Work}
\label{sec:format}

\textbf{Image-to-Video Generation.} Recent advances in diffusion-based generative models have accelerated progress in video synthesis. Works such as AnimateDiff \cite{guo2024animatediff}, Stable Video Diffusion (SVD) \cite{blattmann2023stablevideodiffusion}, and CogVideoX \cite{yang2024cogvideox} demonstrate strong performance in animating still images, while methods like Animate-Anyone \cite{hu2023animateanyone} and ControlNeXt \cite{peng2024controlnext} enable controllable generation through pose conditioning. These I2V models achieve impressive visual fidelity and motion coherence but are highly susceptible to misuse, motivating protective strategies.

\textbf{Adversarial Protection in Generative Models.} Most prior efforts in safeguarding visual content focus on image-based diffusion models. Approaches like AdvDM \cite{liang_advdm}, Mist \cite{liang2023mist},  DiffusionGuard \cite{choi2025diffusionguard}, DCT-Shield \cite{dcts} and PhotoGuard \cite{salman_photoguard} leverage adversarial perturbations to prevent malicious editing, while, Glaze \cite{shan23_glaze} protects against unauthorized editing or style mimicry. 

PRIME \cite{li2024prime} introduces adversarial perturbations to shield videos from malicious editing. To the best of our knowledge, I2VGuard \cite{gui2025i2vguard} is the only prior work that explicitly addresses malicious image-to-video generation from protected images. Importantly, I2VGuard shows that immunization methods designed for image-to-image (I2I) editing do not transfer reliably to image-to-video (I2V) models, highlighting a fundamental gap between adversarial defenses in the two settings. I2VGuard mainly induces spatio-temporal degradation to reduce malicious utility, but degraded outputs can still retain identity cues and residual motion, limiting protection. In contrast, defenses that directly enforce temporal freezing remain largely unexplored.
\section{Preliminaries}
\label{sec:prelim}

\textbf{Diffusion models.} Diffusion models learn to reverse a gradual noising process. Given anoriginal sample $\mathbf{x}_0$, the forward process adds Gaussian noise over timesteps $t=1,\dots,T$, and can be written in closed form as $\mathbf{x}_t = \sqrt{\bar{\alpha}_t}\,\mathbf{x}_0 + \sqrt{1-\bar{\alpha}_t}\,\boldsymbol{\epsilon}$, where $\boldsymbol{\epsilon}\sim\mathcal{N}(0,\mathbf{I})$. A denoising network parameterized by $\theta$ is trained to predict the injected noise (or an equivalent target) from $\mathbf{x}_t$, timestep $t$, and conditioning signals. During generation, a scheduler iteratively applies reverse updates from noisy initialization to obtain a new sample.

\textbf{Video diffusion and I2V generation.} In image-to-video (I2V) systems, diffusion is performed on spatio-temporal latents rather than single images. Let $\mathbf{z}_0\in\mathbb{R}^{T\times C\times H\times W}$ denote the video's latent representation; noising and denoising are applied jointly over all frames, typically with a U-Net or DiT backbone equipped with temporal or spatio-temporal attention. Conditioned on the input image (and optionally text), the model predicts frame-wise noise while enforcing cross-frame consistency through attention. This joint denoising mechanism is central to motion synthesis and therefore provides the key attack surface exploited by Vid-Freeze.

\textbf{Attention Mechanism}. In transformers \cite{vaswani2017attention}, attention maps each token to a weighted combination of value tokens based on query--key similarity. Given token features $\mathbf{X}$, we form queries, keys, and values as $\mathbf{Q}$ $\mathbf{K}$ and $\mathbf{V}$ through learnable projections. Scaled dot-product attention is defined as
\begin{equation}
\mathrm{Attn}(\mathbf{Q},\mathbf{K},\mathbf{V})=\mathrm{softmax}\!\left(\frac{\mathbf{Q}\mathbf{K}^\top}{\sqrt{d_k}}\right)\mathbf{V},
\end{equation}
where $d_k$ is the key dimension. Multi-head attention applies this operation across multiple learned projections and concatenates the resulting heads.

\section{Proposed Method}
\label{sec:method}

\subsection{Threat Model}

    We assume an adversarial setting where a malicious editor employs a pre-trained image-to-video model like CogVideoX or Stable Video Diffusion to perform image-to-video generation. Anticipating this, the defender generates an immunized image by adding near-imperceptible adversarial perturbations to the image using our proposed approach \texttt{Vid-Freeze}. We also assume a white-box setting, where the defender has access to the model and its weights.

\subsection{Motivation}
\label{subsec:motivation}
Prior image immunization approaches for I2V generation often produce visually degraded or spatially inconsistent videos, especially in high-motion regions. However, such distortions do not reliably prevent malicious usage: the generated video can still preserve identity cues and convey harmful intent through residual motion. We therefore propose that a stronger and more meaningful defense is to suppress motion itself, forcing the generated output to remain static across frames.

\textbf{Insights from the denoising process.} First, the denoising network takes concatenated latent representations as input and predicts noise at each sampling step. Consequently, features in early blocks (down blocks in a U-Net or initial blocks in a DiT) remain closer to the input latents, whereas features in deeper blocks are progressively aligned with noise prediction. This also implies that attention in early blocks primarily captures relationships among latent tokens at the current timestep, while attention in deeper blocks increasingly captures relationships in the predicted-noise space.

\begin{figure}[htbp]
    \centering
    \includegraphics[width=\linewidth]{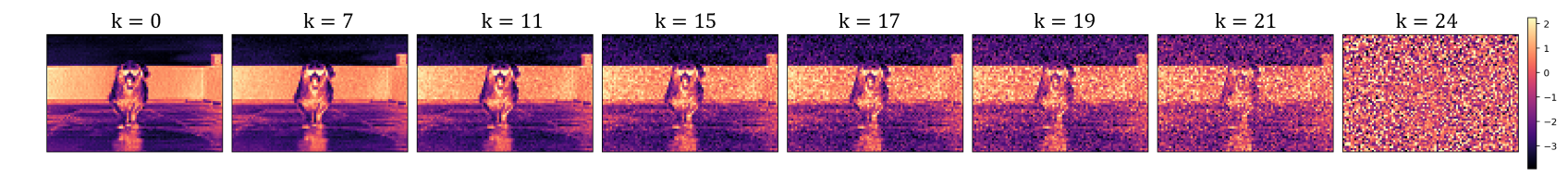}
    \caption{
        Noise predictions across sampling steps ($0\leq k < 25$) for a given frame. Predictions at early sampling steps encode prominent motion-defining structure and progressively become noisier, especially towards the end.
    }
    \Description{Heatmaps of predicted noise for one frame across diffusion steps, showing structured patterns early and increasingly noisy patterns at later steps.}
    \label{fig:temporal-evolution}
\end{figure}

Second, noise predictions exhibit stronger semantic structure at early sampling steps and become increasingly stochastic at later timesteps. This trend follows directly from the DDPM forward noising process, where $q(\mathbf{x}_t\mid\mathbf{x}_{t-1})=\mathcal{N}(\sqrt{1-\beta_t}\,\mathbf{x}_{t-1},\beta_t\mathbf{I})$. Under the standard increasing variance schedule, $\beta_t$ is small at $t=0$ and larger toward $t=T$, so each successive step injects more noise and retains less signal. This behavior is also evident in Fig.~\ref{fig:temporal-evolution}, where we visualize the predicted noise at each sampling step for CogVideo. Together, these observations motivate the following hypothesis.

\textbf{Hypothesis}: Based on these observations, we hypothesize that if the predicted noise corresponding to all frames during inference is aligned with that of the first frame, then the final generated frames will converge to the first frame. To realize this in practice, we further hypothesize that biasing attention in deeper blocks toward the first frame can drive successive frames to collapse onto it, thereby inducing temporal freezing.

\begin{figure}[htbp]
    \centering
    \includegraphics[width=\linewidth,]{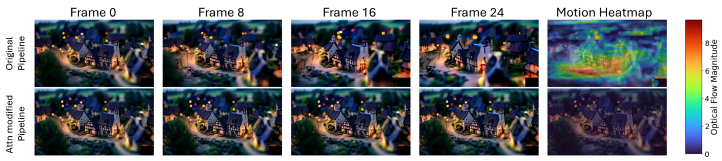}
    \caption{
        Sanity check of the hypothesis. Top rows show motion under the standard pipeline. Bottom rows show inference with temporal attention hard-coded so that first-column entries are set to 1, yielding static videos.
    }
    \Description{Comparison between normal inference and hard-coded temporal attention, where forcing first-frame attention yields frozen outputs.}
    \label{fig:left-heavy-validation}
\end{figure}

We further perform a sanity check for this hypothesis by explicitly manipulating temporal attention in the up-blocks of SVD's U-Net. Specifically, we overwrite the temporal attention matrix (post soft-max) by setting the first column to 1 and all remaining columns to 0. As shown in Fig.~\ref{fig:left-heavy-validation}, this intervention consistently yields frozen or near-static videos across diverse inputs. These results motivate us to propose adversarial defense methods aimed at manipulating attention, as described.

\subsection{Problem Formulation}

Different image-to-video architectures realize temporal (across frames) interactions in different ways. For example, SVD employs decoupled spatial and temporal attention layers within its U-Net, whereas CogVideoX uses spatio-temporally coupled attention because its DiT processes a sequence of tokens spanning both patches and frames. Because our method is attention-based, a single freezing objective is not equally effective for both models. We therefore instantiate the same high-level goal---suppressing motion through attention manipulation---using two architecture-specific formulations. The following subsections describe the strategies for decoupled temporal attention (SVD) and coupled spatio-temporal attention (CogVideoX).

\textbf{Decoupled Temporal Attention (SVD).} In architectures such as Stable Video Diffusion (SVD), temporal and spatial attention are decoupled. 
Temporal attention operates across frames independently for each spatial token. 
Let $F$ denote the number of frames, $S$ the number of spatial tokens per frame, $H$ the number of attention heads, and $L$ the number of temporal attention layers. 
For each layer $\ell$, the temporal attention-score tensor can therefore be written as
\begin{equation}
A_{\text{temp}}^{(\ell)} \in \mathbb{R}^{S \times H \times F \times F},
\end{equation}
where each spatial token and attention head produces an $F \times F$ attention matrix capturing cross-frame dependencies. 
The attention matrix is obtained through a softmax operation, so each entry satisfies
\begin{equation}
0 \le A_{\text{temp}}^{(\ell,s,h)}(i,j) \le 1, \qquad
\sum_{j=1}^{F} A_{\text{temp}}^{(\ell,s,h)}(i,j) = 1.
\end{equation}

Temporal attention directly controls interactions between frames. Freezing can therefore be achieved by redirecting attention toward the first frame. Therefore, we define a target attention pattern
\begin{equation}
A_{tgt} =
\begin{bmatrix}
1 & 0 & \cdots & 0 \\
1 & 0 & \cdots & 0 \\
\vdots & \vdots & \ddots & \vdots \\
1 & 0 & \cdots & 0
\end{bmatrix},
\end{equation}
which encourages all frames to attend to the first frame. When multiplied by the value tensor, this attention pattern causes each frame's resultant representation to be dominated by the first frame. 
The freezing loss is defined as
\begin{equation}
\mathcal{L}_{freeze} =
\frac{1}{L}\sum_{\ell=1}^{L}
\left\|A_{\text{temp}}^{(\ell)} - A_{tgt}\right\|_F^2,
\end{equation}
where $A_{\text{temp}}^{(\ell)} \in \mathbb{R}^{S \times H \times F \times F}$ and $A_{tgt} \in \mathbb{R}^{F \times F}$ is broadcast over the $S$ and $H$ dimensions.

\textbf{Coupled Spatio-Temporal Attention (CogVideoX).} Unlike SVD, CogVideoX models spatial and temporal interactions jointly within a single attention mechanism. 
At layer $\ell$, the attention tensor can be written as
\begin{equation}
A^{(\ell)} \in \mathbb{R}^{H \times (FS) \times (FS)},
\end{equation}
where each attention head produces a $(FS) \times (FS)$ matrix capturing interactions between all spatial tokens across all frames.

\begin{figure*}[t]
\centering
\includegraphics[width=\textwidth]{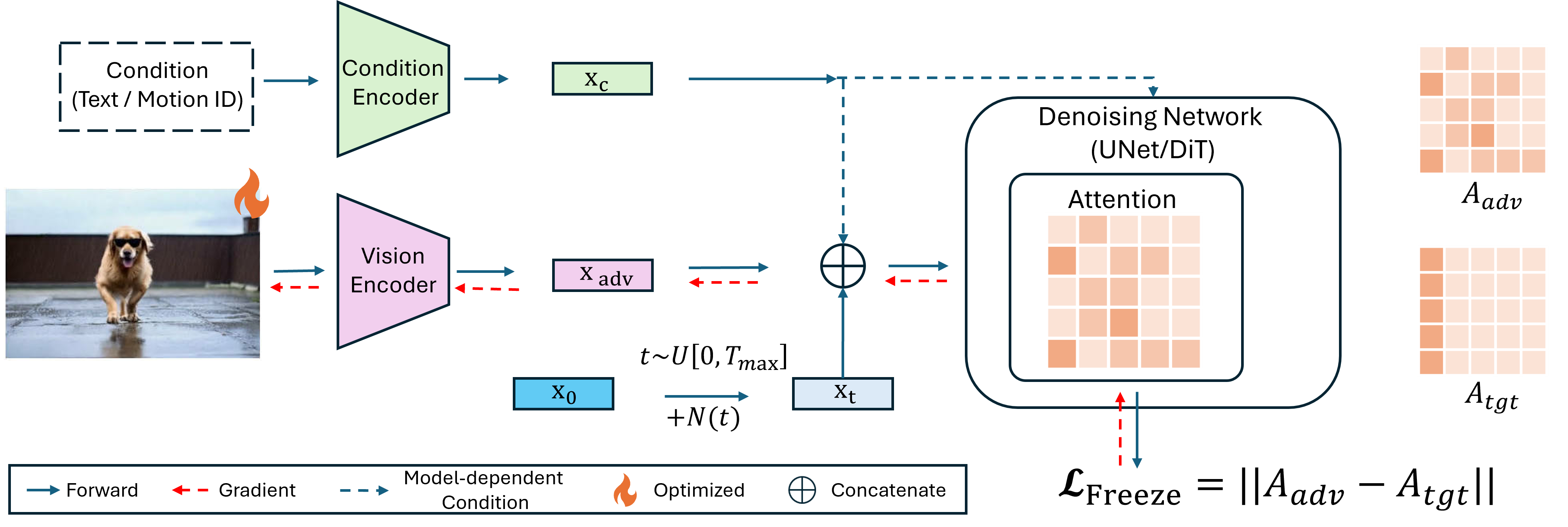} 
    \caption{\textbf{Overview of Vid-Freeze.} Given an input image, Vid-Freeze optimizes a small adversarial perturbation under a bounded pixel budget so that the perturbed image drives the I2V denoiser toward a temporal-freezing behavior. During optimization, we extract intermediate attention maps and apply a freezing objective $\mathcal{L}_{freeze}$ that suppresses cross-frame motion propagation, causing later frames to collapse toward the first frame.}
    \Description{Method diagram showing input image perturbation optimization with attention-based freezing loss to drive generated video frames toward the first frame.}
    \label{fig:method}
\end{figure*}
In this setting, cross-frame attention is distributed across many spatial tokens, making it difficult to isolate temporal interactions using simple frame-level constraints. 
Empirically, using an attention target with higher weights in the first temporal column span (columns coresponding to the first frame) is less effective for suppressing motion and will be further discussed in the supplementary material.
Instead, we define the freezing loss by reducing the overall magnitude of the attention tensors across layers:
\begin{equation}
\mathcal{L}_{freeze} =
\frac{1}{L}\sum_{\ell=1}^{L}
\left\|A^{(\ell)}\right\|_F^2,
\end{equation}
where $A^{(\ell)} \in \mathbb{R}^{H \times (FS) \times (FS)}$ denotes the attention tensor at layer $\ell$.

\subsection{Expectation over Generation Conditions}

The freezing loss defined in the previous sections is evaluated under different generation conditions of the video diffusion models. 
To obtain perturbations that generalize across these conditions, we minimize the expected freezing loss over both the model conditioning variables and the diffusion timesteps.

Formally, the overall optimization objective is
\begin{equation}
\min_{\|\delta\|_\infty \le \epsilon}
\mathbb{E}_{c,t}
\left[
\mathcal{L}_{freeze}(x_0+\delta; c, t)
\right],
\label{eq:objective}
\end{equation}
where $c$ denotes the set of model-specific conditioning variables and $t$ denotes the diffusion timestep sampled during the denoising process. Fig. \ref{fig:method} shows an overview of our proposed method. 
The conditioning variables differ across the evaluated models. 
For Stable Video Diffusion (SVD), the conditioning variables include frames-per-second (fps), and the motion-id parameter that controls motion strength. 
For CogVideoX, the conditioning variable corresponds to the text prompt.

In our experiments, the expectation over diffusion timesteps is approximated by sampling a subset of timesteps during optimization. 
For CogVideoX, we do not perform expectation over prompts. Instead, we either use a null prompt during optimization or a caption of the image as a prompt.

\section{Experiments}
\label{sec:experiments}
\textbf{Data and Metrics}.
To evaluate our method, we curate a dataset tailored for SVD and CogVideoX. We begin with a pool of 300 images collected from the web. For each image, we generate videos using both models and examine the resulting outputs. We filter out samples that produce low-quality generations or exhibit unstable motion, retaining only images that lead to visually coherent videos with natural motion in the setting. From this filtered pool, we construct two evaluation sets of 100 images each for SVD and CogVideoX. The final datasets contain diverse subjects and scenes, including humans, animals, objects, and natural environments. We evaluate our method using perceptual, motion, and quality metrics. LPIPS \cite{LPIPS} measures the imperceptibility of perturbations, while motion is assessed via average dense optical flow magnitude and inter-frame SSIM difference \cite{ssim}. We also use three VBench metrics \cite{huang2023vbench}—Subject Consistency, Aesthetic Quality, and Image Quality—to assess overall generation quality. 
We also conduct a human evaluation of protection effectiveness. Evaluators rate protection quality on a 0--5 scale based on videos generated from immunized images. A detailed description of the human-evaluation setup is provided in the supplementary material.

\textbf{Implementation Details}. Unless otherwise specified, adversarial optimization is performed with a pixel perturbation budget of $\epsilon=16$ for 1000 iterations. We optimize Vid-Freeze using the objective in Eq.~\ref{eq:objective} and projected gradient descent (PGD) \cite{pgd}. Pseudocode is provided in the supplementary material. We compare against I2VGuard as a baseline. Because the official I2VGuard implementation is unavailable, we re-implement it and additionally report direct comparisons against the qualitative results in the original paper (see supplementary material) to support fair evaluation.

We evaluate on CogVideoX-2B and Stable Video Diffusion (SVD), two widely used image-to-video models that allow adversarial optimization within a practical compute budget (~ 80 GB GPU memory). Since optimization backpropagates through the denoising network, memory usage is high due to large intermediate activations. Therefore, during immunization, we limit the number of frames used for optimization to at most 9 for CogVideoX and 16 for SVD.

\textbf{Motion characteristics of the evaluated models.} The evaluated video diffusion models exhibit distinct motion profiles. Stable Video Diffusion (SVD) provides explicit control over motion strength via the motion bucket (motion-id) parameter, which modulates motion magnitude and often produces pronounced global dynamics, including camera motion (e.g., panning and zooming), in addition to subject motion. In contrast, CogVideoX-2B is primarily guided by text prompts and does not expose an explicit motion-strength control. Empirically, CogVideoX-2B tends to produce more localized, subject-centric motion, with less pronounced global motion than SVD. These differences yield distinct baseline motion distributions across models and are accounted for when evaluating freezing behavior.

\section{Results}
\label{sec:results}

\begin{table*}[t]
\centering
\resizebox{0.85\textwidth}{!}{

\begin{tabular}{l|l|cc|cc|c|c}
\hline
\textbf{Method} & \textbf{Model} & \shortstack[c]{Temporal \\ SSIM ($\downarrow$)} & \shortstack[c]{Flow \\ Mag. ($\downarrow$)} & \shortstack[c]{Aesthetic} & \shortstack[c]{Subject \\ Consistency} & \shortstack[c]{LPIPS \\ $\mathbf{X},\mathbf{X_{adv}}$ ($\downarrow$)} & \shortstack[c]{Human \\ Eval. ($\uparrow$)} \\
\hline
Original Image &  & 0.077 & 0.647 & 5.401 & 0.990 & -- & -- \\
 I2VGuard* & CogVideoX& 0.200 & 2.480 & 3.662 & 0.961 & 0.335 & 2.9 \\
 VidFreeze &  & \textbf{0.006} & \textbf{0.050} & 4.990 & 0.989 & \textbf{0.282} & \textbf{4.9} \\
\hline
Original Image &  & 0.305 & 4.096 & 4.350 & 0.980 & -- & -- \\
I2VGuard* & SVD & 0.186 & 2.246 & 3.712 & 0.974 & \textbf{0.213} & 3.4 \\
VidFreeze &  & \textbf{0.014} & \textbf{0.168} & 4.301 & 0.990 & 0.223 & \textbf{4.5} \\ 
\hline
\end{tabular}
}
\caption{\textbf{Quantitative comparison of attack methods for image immunization on CogVideoX I2V and SVD pipelines.} We report mean values for all metrics. Human ratings (/5) are shown in the last column. All results use $\epsilon = 16$. Near-zero flow magnitude and temporal SSIM show Vid-Freeze's strong motion-blocking ability. (* denotes our self-implemented I2VGuard baseline, since the official code is not publicly available.)}
\label{tab:vid_metrics}
\end{table*}

\begin{figure*}[t]
\centering
\includegraphics[width=\textwidth]{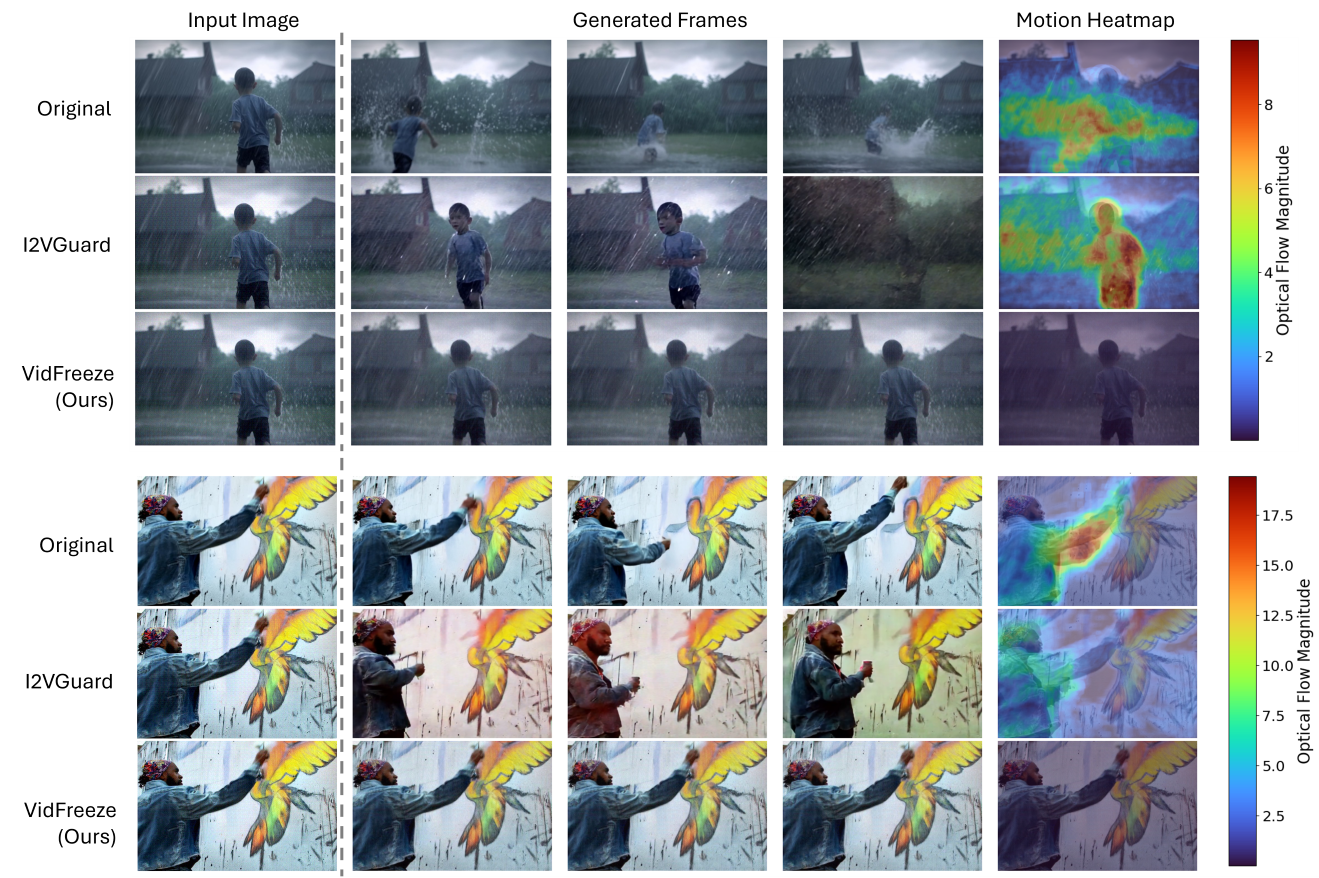}
\caption{Qualitative results comparing immunization strategies on CogVideoX. Vid-Freeze produces frozen videos with near-identical frames, whereas I2VGuard produces a spatio-temporally degraded video that may still harm victims.}
\Description{Frame grids for CogVideoX comparing original, I2VGuard, and Vid-Freeze outputs, where Vid-Freeze remains nearly static across frames.}
\label{fig:qual_results_cog}
\end{figure*}
\subsection{Immunization Quality}

\begin{figure*}[t]
\centering
\includegraphics[width=\textwidth]{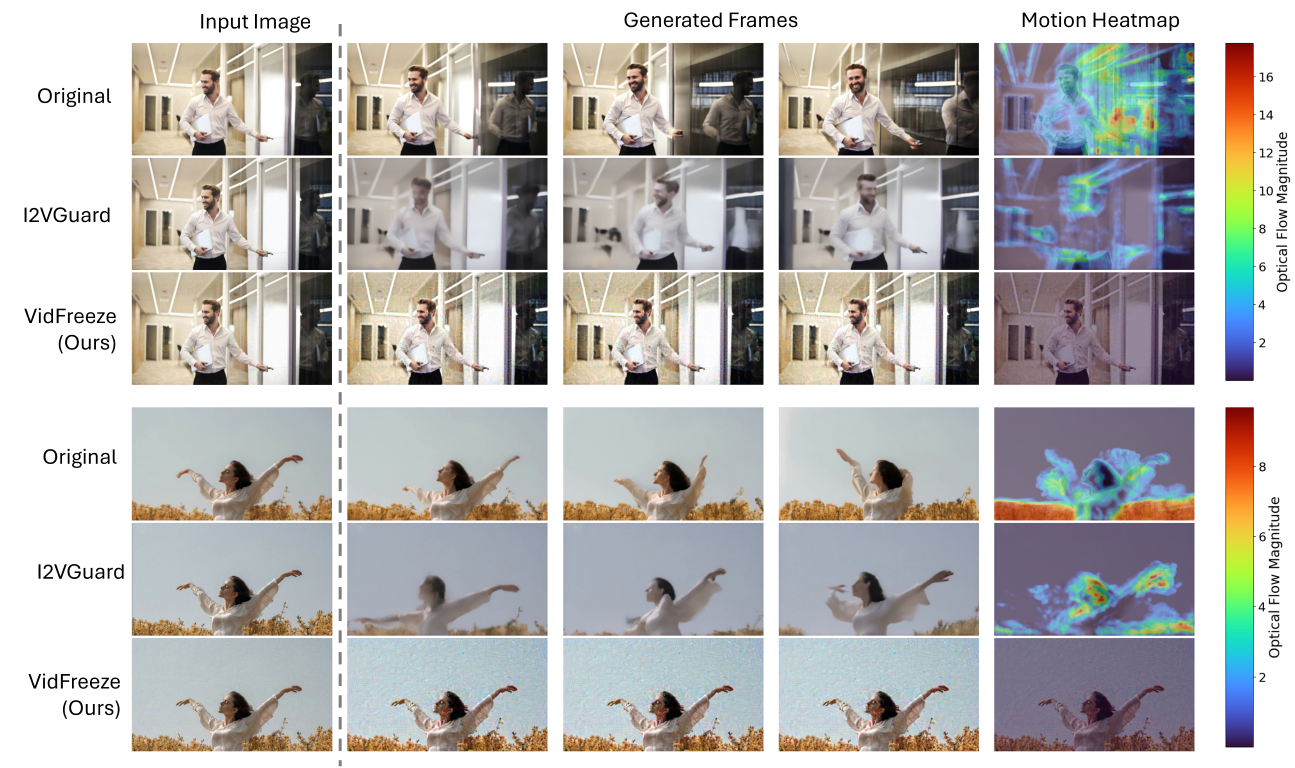}
\caption{Qualitative results comparing immunization strategies on SVD. Vid-Freeze produces frozen videos with near-identical frames, whereas I2VGuard produces a spatio-temporally degraded video that may still harm victims.}
\Description{Frame grids for SVD comparing original, I2VGuard, and Vid-Freeze outputs, where Vid-Freeze keeps temporal content nearly unchanged.}
\label{fig:qual_results_svd}
\end{figure*}

\subsubsection{Qualitative Results}
We compare videos from original images with those generated from images immunized by I2VGuard (our implementation) and Vid-Freeze across SVD and CogVideoX-2B. As shown in Fig.~\ref{fig:qual_results_cog} and Fig.~\ref{fig:qual_results_svd}, original images yield smooth, coherent motion (first row), whereas I2VGuard introduces visible spatial artifacts across frames (second row). Notably, despite these distortions (e.g., in hands and faces), the prompt intent often remains recognizable, and subject identity is sometimes largely preserved, which limits practical protection. In contrast, Vid-Freeze produces static (frozen) videos with minimal temporal variation and provides stronger protection against malicious animation. We observe the same trend for both CogVideoX-2B and SVD. Additional qualitative results, including direct comparisons with those reported in the I2VGuard paper, are provided in the supplementary material.

\subsubsection{Quantitative Results}
Table \ref{tab:vid_metrics} shows that Vid-Freeze attains the extremely low mean $\Delta$SSIM and optical flow magnitude, confirming that the generated videos remain practically static across frames. These results indicate that Vid-Freeze achieves temporal freezing across both evaluated models, each with different architectures and motion charactersitics. This validates the effectiveness of the proposed freezing attack in halting motion and immunizing images against malicious video generation. 
We also note that Vid-Freeze yields slightly higher motion metrics on SVD than on CogVideoX. This is because videos generated from immunized images via SVD can sometimes exhibit global flickering artifacts (e.g., frame-to-frame brightness fluctuations or motion in spatial artifacts). However, these effects do not correspond to meaningful subject or camera motion, and the videos remain semantically frozen. We provide representative examples in the supplementary material.
It is also worth noting that Vid-Freeze provides protection through temporal freezing rather than spatial degradation. Therefore, it is not expected to achieve the 'best' scores (lower is better for spatial degradation) on metrics such as aesthetics or subject consistency, which primarily assess spatial integrity. 
Human-evaluation scores also demonstrate a clear preference for Vid-Freeze.

\subsection{Attention Visualization}
\begin{figure}[htbp]
    \centering
    \includegraphics[width=\linewidth]{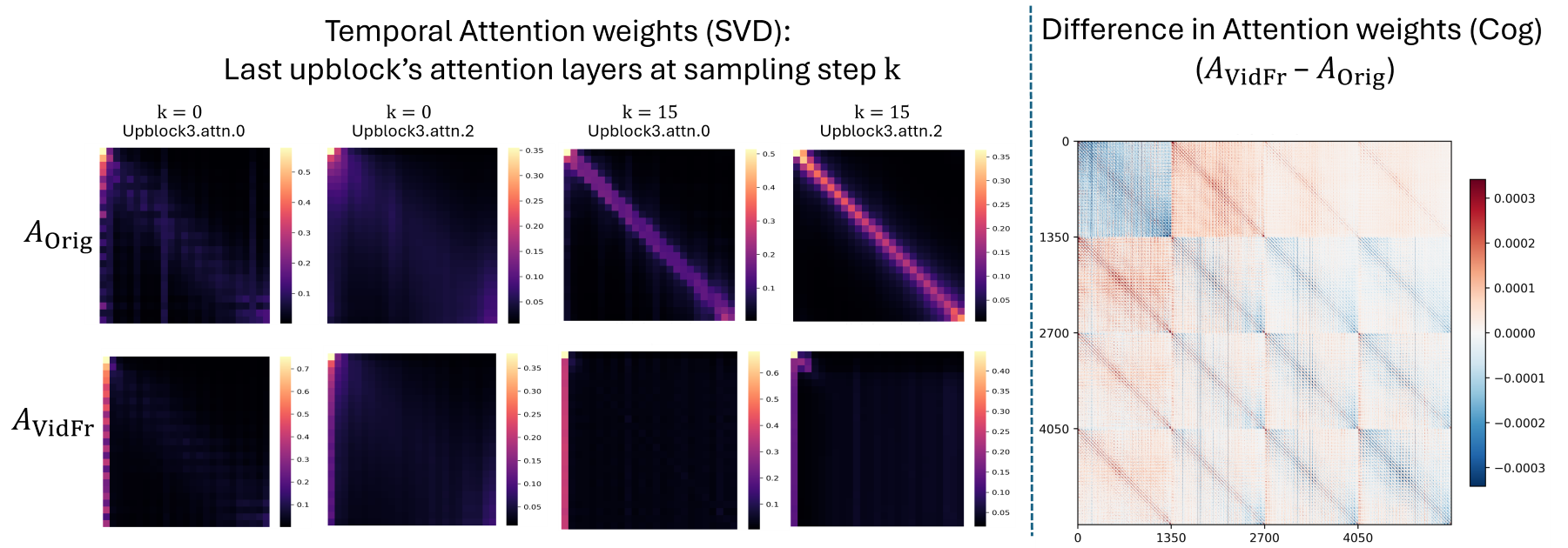}
    \caption{
        Attention visualization for SVD and CogVideoX.
        For SVD, we show temporal attention matrices from upblocks across different sampling steps,
        comparing original and frozen generations. For CogVideoX, we show attention-difference heatmap
        (immunized minus original), where red and blue denote increased and decreased weights,
        respectively. Attention shifts toward the first frame after immunization with VidFreeze.
    }
    \Description{Attention maps showing stronger first-frame focus after immunization for both SVD temporal attention and CogVideoX attention differences.}
    \label{fig:attention-heatmaps}
\end{figure}

In Stable Video Diffusion (SVD), spatial and temporal attention are factorized into separate modules, allowing us to inspect temporal dynamics in isolation. We visualize temporal attention matrices (averaged across heads and spatial tokens) from the up-blocks at the first ($k=0$) and an intermediate ($k=15$) sampling step, and compare original generations with Vid-Freeze generations. As shown in Fig.~\ref{fig:attention-heatmaps} (left panel), original videos exhibit strong temporal interactions across neighboring frames. The broadened diagonal pattern indicates high similarity between adjacent frames, which decreases as frame separation increases. By contrast, frozen videos show attention that is much more concentrated in the first column and more uniformly distributed elsewhere, indicating that features across frames are nearly identical to those of the first frame.

For CogVideoX, the denoiser is a DiT and uses unified spatio-temporal self-attention, so the attention matrix jointly captures spatial and temporal dependencies. Because the attention matrix is large and individual entries are numerically small, direct visualization of original vs. immunized is less informative; therefore, we visualize the difference between the two attention matrices instead to hightlight the shift in attention weights. In Fig.~\ref{fig:attention-heatmaps}, the figure on the right shows $A_{VidFreeze} - A_{Original}$, where red and blue denote the increase and decrease in attention weights after immunization, respectively. Note that the block-like pattern arises because tokens are grouped by frame: each contiguous token group contains all patch tokens from one frame, so frame-wise interactions appear as matrix blocks. The difference heatmap clearly shows a shift of attention mass toward the first frame, consistent with the temporal-freezing behavior induced by Vid-Freeze. We also note that we only show interactions between vision tokens (and not text tokens) for clear viusalization.

\subsection{Evolution of Frames across Denoising Timesteps}

\begin{figure}[htbp]
    \centering
    \includegraphics[width=0.95\columnwidth]{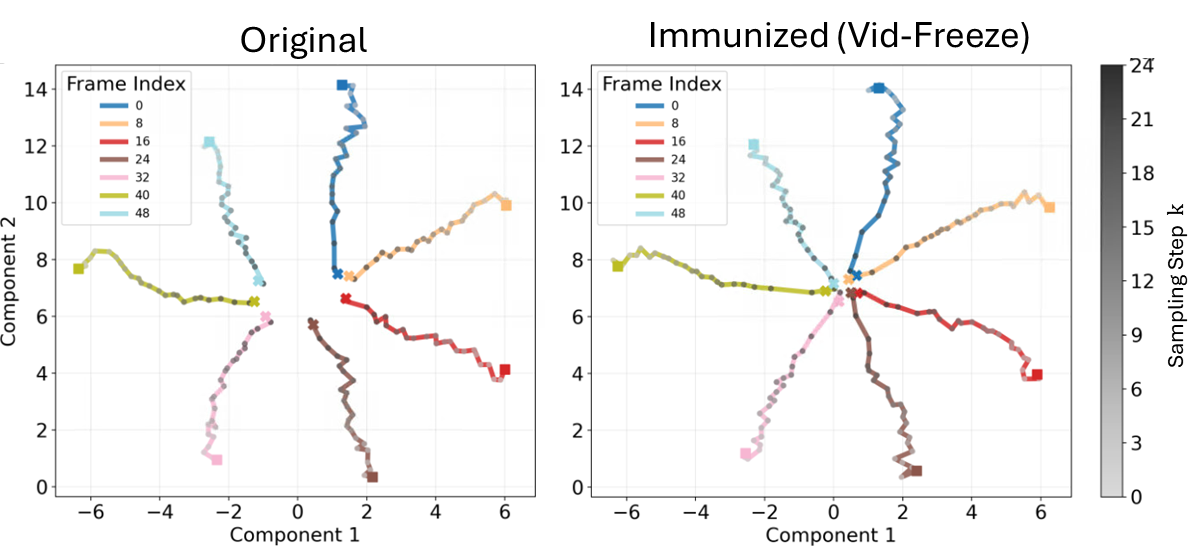}
    \caption{
        UMAP projection of frame latents across sampling steps $0\leq k<25$ for original and immunized inputs. Each trajectory starts at a square marker and ends at a cross marker. For the original input, final latents from different frames remain separated. For the Vid-Freeze-immunized input, final latents from different frames converge to a shared region in latent space, consistent with frozen-video behavior.
    }
    \Description{Two-dimensional latent trajectories across timesteps showing frame endpoints separated for original input and clustered for Vid-Freeze-immunized input.}
    \label{fig:tsne_latent_evolution}
\end{figure}

To further analyze temporal freezing, we project frame-wise latents across denoising steps with UMAP \cite{mcinnes2018umap} (Fig.~\ref{fig:tsne_latent_evolution}), which maps high-dimensional latents to 2D while preserving local neighborhood structure. Circles denote initial latents, crosses denote final latents, and colored points denote intermediate denoising states. For original inputs, final frame latents remain well separated, indicating diverse dynamics. For immunized inputs, final latents from all frames converge to a common region, showing near-identical frame representations and therefore temporally frozen videos

\subsection{Ablations}

To evaluate the contribution of different components in our framework, we conducted a series of ablation studies. 

\subsubsection{Sampling Steps} 

\begin{figure}[t]
    \centering
    \includegraphics[width=\linewidth]{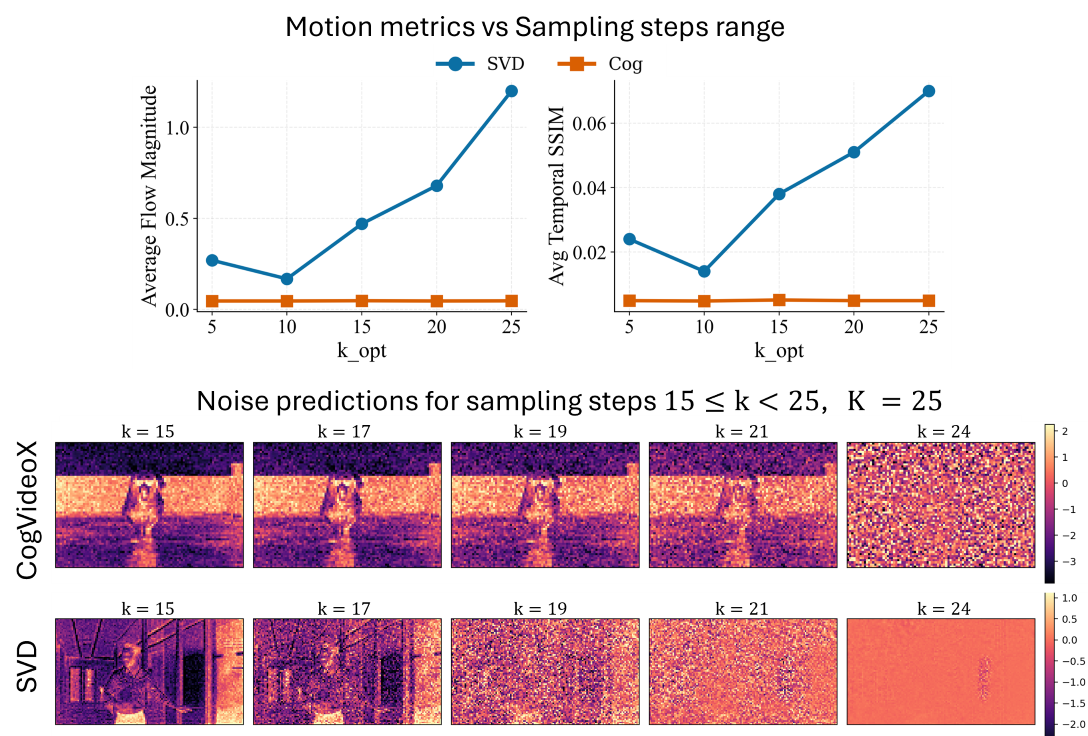}
    \caption{
        Ablation of sampling steps used during immunization. Left panel shows the effect of using first $k$ sampling steps on motion freezing. Right panel shows noise predictions for a fixed frame across sampling steps $15 \leq k < 25$, showing progressively diminishing semantic structure and increasing noise.
    }
    \Description{Ablation plots showing freezing quality versus optimization timestep range and example late-step noise predictions becoming progressively less structured.}
    \label{fig:timestep_ablation}
\end{figure}

As motivated in Sec.~\ref{subsec:motivation}, noise predictions at smaller $k$ values retain more structure. Consequently, attention matrices in deeper layers at these steps provide more informative gradients for immunization. We therefore ablate $k_{\mathrm{opt}}$ to identify the best sampling range $[0, k_{\mathrm{opt}})$. While the total number of sampling steps for inference and optimization is fixed at $K=25$, each PGD step samples $k$ from $[0, k_{\mathrm{opt}})$, where $k_{\mathrm{opt}} \leq K$. As shown in the top panel of Fig.~\ref{fig:timestep_ablation}, CogVideoX performs consistently well across $k_{\mathrm{opt}}$ values, indicating limited sensitivity to the timestep range. This is supported by the bottom-panel visualization, where noise predictions remain relatively structured even at late timesteps. In contrast, SVD improves as $k_{\mathrm{opt}}$ increases to about 10, then degrades for larger $k_{\mathrm{opt}}$, consistent with significantly noisier late-step predictions. Qualitative results for this ablation are provided in the supplementary material.

\subsubsection{Pixel Budget}

\begin{figure*}[t!]
    \centering
    \includegraphics[width=\linewidth]{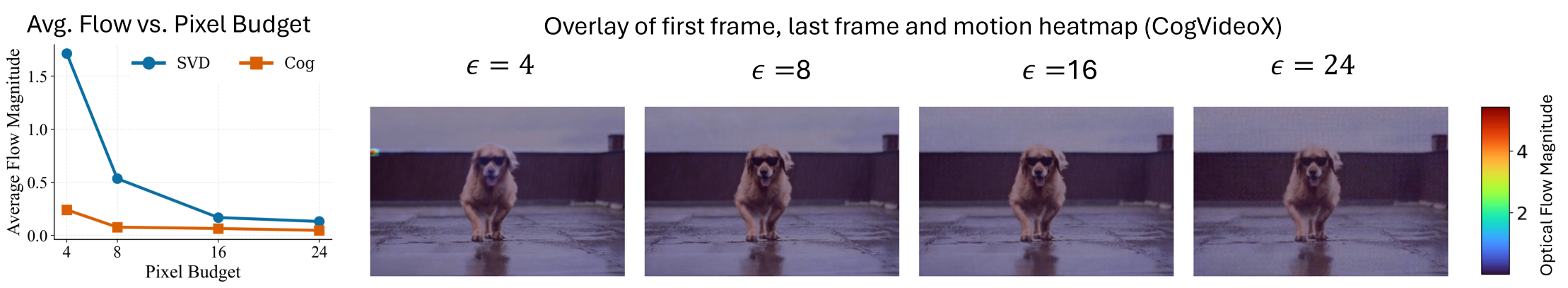}
    \caption{
            Ablation over pixel budgets. The left panel shows that CogVideoX and SVD exhibit strong temporal freezing for $\epsilon \in [4,24]$ and $\epsilon \in [8,24]$, respectively. The right panel presents qualitative results for CogVideoX, showing overlays of the first and last frames with optical-flow magnitude heatmaps across different pixel budgets, clearly indicating frozen videos.
    }
    \Description{Pixel-budget ablation combining quantitative freezing curves and qualitative overlays with optical-flow heatmaps across perturbation strengths.}
    \label{fig:flow_mag_vs_pixel budget.}
\end{figure*}

We ablate the perturbation budget to analyse the sensitivity of immunization to changes in pixel budget. Figure~\ref{fig:flow_mag_vs_pixel budget.} shows that the average flow magnitude increases as the pixel budget is reduced from 24 to 4. For CogVideoX, $\epsilon=4$ is sufficient to freeze motion, and a significant fraction of images are successfully immunized even at very low budgets (down to 2 pixels). In contrast, SVD is less reliable at very low budgets because its Gaussian noising stage acts as an implicit purification step that can attenuate adversarial perturbations. As a result, freezing in SVD is inconsistent at the smallest budgets, but becomes stable at $\epsilon=8$, and remaining well within commonly used perturbation ranges in the adversarial-attack literature. We also present qualitative CogVideoX results in Fig.~\ref{fig:flow_mag_vs_pixel budget.} (right panel), showing overlays of the first frame, last frame, and optical-flow heatmap across pixel budgets. These visualizations confirm that the generated videos are temporally frozen.

\subsubsection{Motion generating conditions}
We ablate over motion generating conditions described in Sec. \ref{sec:experiments}, to assess their effect on immunization quality.
For CogVideoX, we do not optimize an expectation over text prompts, since defining and sampling a representative prompt distribution for each image is open-ended and computationally expensive. Instead, we compare optimization with a null prompt versus an image caption. Because captions provide semantic context, they are more likely to overlap with unknown malicious prompts. As shown in Table~\ref{tab:caption-ablation}, caption conditioning lowers both $\Delta\text{SSIM}$ and flow magnitude. Both settings provide strong protection: the null prompt is simpler, while caption conditioning offers the strongest freezing.
For SVD, we analyze the effect of using Expectation over Generation Conditions (EoGC) during optimization. Table~\ref{tab:caption-ablation} shows that using EoGC is necessary to acheive meaningful immunization with consistent freezing.

\begin{table}[htbp]
\centering
\setlength{\tabcolsep}{10.0pt} 
\begin{tabular}{lcc}
\hline
\textbf{Condition type (Cog)} & $\Delta$SSIM ($\downarrow$) & Flow Mag. ($\downarrow$) \\
\hline
Null Prompt  & 0.0068 & 0.0601 \\
Img. Caption & \textbf{0.0056} & \textbf{0.0497} \\
\hline\hline
\textbf{Condition type (SVD)} & & \\
\hline
w/o EoGC & 0.165 & 1.81 \\
w EoGC   & \textbf{0.014} & \textbf{0.168} \\

\hline
\end{tabular}
\caption{Ablation study on the effect of motion generating conditions used during optimization}
\label{tab:caption-ablation}
\end{table}

\subsection{Robustness to Common Purification Methods}
\begin{figure}[htbp]
    \centering
    \includegraphics[width=\linewidth]{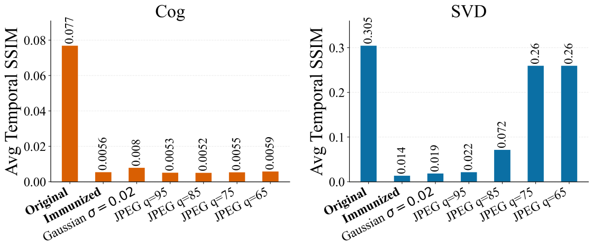}
    \caption{Robustness of Vid-Freeze immunization under Gaussian noise and JPEG recompression for CogVideoX and SVD. Lower values indicate stronger temporal freezing.}
    \Description{Robustness plot comparing original and immunized images under Gaussian noise and JPEG compression across CogVideoXand SVD metrics.}
    \label{fig:robustness_common_purification}
\end{figure}

We further evaluate the robustness of immunized images under common image-space transformations encountered in real-world pipelines. Specifically, we consider JPEG recompression and additive Gaussian noise, both of which are known to weaken adversarial perturbations. Our results in Fig. \ref{fig:robustness_common_purification} show that the proposed immunization remains effective under Gaussian noise with $\sigma = 0.02$, consistently producing stand-still or near-static videos. Under JPEG recompression, robustness differs across backbones: for CogVideoX, protection remains strong down to 65\% quality, whereas for SVD it is reliable down to 85\% quality and degrades at lower qualities. Improving robustness to stronger purification remains an important direction for future work, and methods such as \cite{Chen_editshield}, \cite{dcts}, \cite{honig2024adversarial} etc provide a promising starting point.

\section{Conclusion}
\label{sec:conclusion}

We presented Vid-Freeze, a proactive defense against malicious image-to-video generation that protects images by enforcing temporal freezing rather than relying on spatial degradation. This provides a more meaningful form of protection, since malicious intent may still be conveyed by spatially degraded outputs, potentially harming innocent individuals. By targeting attention dynamics during adversarial optimization, Vid-Freeze suppresses motion synthesis. Across CogVideoX and SVD, the method consistently reduces motion metrics, produces near-static outputs under diverse prompts, and remains effective under common purification transforms. Human evaluation further supports the qualitative advantage of Vid-Freeze over prior immunization strategies, indicating stronger practical protection against misuse.

\textbf{Limitations.}
Our approach is primarily studied in the white-box setting and requires access to model internals, including attention matrices and intermediate representations. This limits direct applicability to closed-source systems. At the same time, our results establish temporal freezing as a meaningful protection objective and provide a strong foundation for broader defenses. We hope this motivates future research on transferable and black-box freezing-based protection methods that generalize across architectures, training regimes, and deployment conditions while preserving strong protection.



\bibliographystyle{ACM-Reference-Format}
\bibliography{main} 

\end{document}